\documentclass[a4paper, 10pt, conference]{ieeeconf}

\newif\ifarwfinalcopy

\arwfinalcopytrue  

\IEEEoverridecommandlockouts

\usepackage{graphics} 
\usepackage{epsfig} 
\usepackage{mathptmx} 
\usepackage{times} 
\usepackage{amsmath} 
\usepackage{amssymb}  

\usepackage{lineno}
\usepackage{tikzpagenodes}
\usepackage{background}

\usepackage{siunitx} 

\ifarwfinalcopy
\backgroundsetup{color=white}
\else

\linenumbers

\newcommand{\MyARWConfidentialLogo}{
\begin{tikzpicture}[remember picture,overlay]
\node[align=center,text=blue] at ([yshift=1em]current page text area.north) {\Large \#\#\# ARW 2021 SUBMISSION: CONFIDENTIAL REVIEW COPY \#\#\#};
\end{tikzpicture}%
}

\SetBgContents{\MyARWConfidentialLogo}
\SetBgPosition{current page.north west}
\SetBgOpacity{0.5}
\SetBgAngle{0.0}
\SetBgScale{1.0}

\fi

\title{\LARGE \bf Continuous Target-free Extrinsic Calibration of a Multi-Sensor System \\from a Sequence of Static Viewpoints}

\ifarwfinalcopy
\author{Philipp Glira$^{1}$, Christoph Weidinger$^{1}$, Johann Weichselbaum$^{1}$
\thanks{*The research leading to these results has received funding from the Mobility of the Future programme. Mobility of the Future is a research, technology and innovation funding programme of the Republic of Austria, Ministry of Climate Action. The Austrian Research Promotion Agency (FFG) has been authorised for the programme management.}
\thanks{$^{1}$AIT Austrian Institute of Technology,
AAS Assistive and Autonomous Systems, 1210 Vienna, Austria {\tt\small philipp.glira@ait.ac.at}, {\tt\small christoph.weidinger@ait.ac.at}, {\tt\small johann.weichselbaum@ait.ac.at}}%
}
\else
\author{Anonymous, J.D.}
\fi

\begin{document}

\maketitle


\begin{abstract}

Mobile robotic applications need precise information about the geometric position of the individual sensors on the platform. This information is given by the extrinsic calibration parameters which define how the sensor is rotated and translated with respect to a fixed reference coordinate system. Erroneous calibration parameters have a negative impact on typical robotic estimation tasks, e.g.\ SLAM. In this work we propose a new method for a continuous estimation of the calibration parameters during operation of the robot. The parameter estimation is based on the matching of point clouds which are acquired by the sensors from multiple static viewpoints. Consequently, our method does not need any special calibration targets and is applicable to any sensor whose measurements can be converted to point clouds.  We demonstrate the suitability of our method by calibrating a multi-sensor system composed by 2 lidar sensors, 3 cameras, and an imaging radar sensor.

\end{abstract}

\section{INTRODUCTION}

\begin{figure}[thpb]
	\centering
	\includegraphics[width=\linewidth]{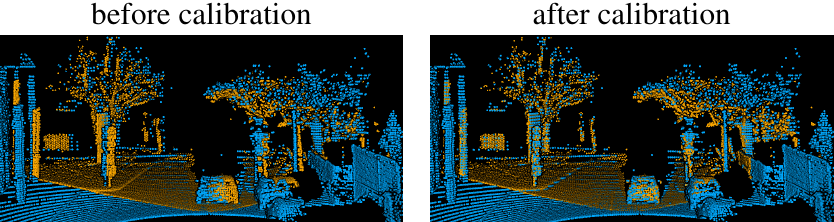}
	\caption{Effect of an extrinsic sensor calibration on the relative orientation of two lidar point clouds.}
	\label{fig:extrinsic_calibration}
\end{figure}

Robots are typically equipped with several sensors to continuously observe their surroundings. For this purpose, various sensor modalities are used due to the specific strengths and weaknesses of each modality. The most commonly used sensors for environment perception are cameras, lidar sensors, radar sensors, ultrasound sensors, and infrared sensor. The data of these sensors is combined by means of \textit{sensor fusion} to get a more complete, accurate, and reliable description of the environment. However, in order to properly fuse the different sensor data streams, they need to be correctly aligned. A misalignment exists if \textit{systematic} discrepancies between the data of different sensors are observed. One way to minimize such discrepancies is a proper \textit{extrinsic calibration}\footnote{sometimes also denoted as \textit{mounting calibration}} of the sensors (Fig.~\ref{fig:extrinsic_calibration}).

The extrinsic calibration defines the rotational and translational offset w.r.t.\ to a fixed reference coordinate system. It has 6 degrees of freedom (DoF) and is  typically defined by 3 Euler angles\footnote{or by an equivalent quaternion} ($\alpha_x$, $\alpha_y$, $\alpha_z$) and 3 translation components ($t_x$, $t_y$, $t_z$) \cite{siegwart2011a}.

We propose in this work a new online target-free calibration method. The main advantages in comparison to previous works are that:
\begin{itemize}
	\item the calibration is widely sensor agnostic, as it is applicable to any sensor whose measurements can be converted to point clouds
	\item the calibration parameter estimates are continuously improved during the operation of the robot until a user-defined threshold is reached
	\item multiple sensors can be calibrated at the same time.
\end{itemize}

The rest of the paper is organized as follows. Related work is reviewed in the subsequent section~\ref{sec:related_work}. The calibration design, procedure, and estimation are described in sections \ref{sec:calibration_design} to \ref{sec:calibration_estimation}. The mathematical description of the optimization task is given in section~\ref{sec:optimization}. A few details about the implementation are reported in section~\ref{sec:implementation}. Finally, experimental calibration results from an 8 minute drive are summarized in section~\ref{sec:experimental_results}.

\section{RELATED WORK}
\label{sec:related_work}

A variety of extrinsic calibration methods have been published over the last few decades -- an overview was recently published by \cite{nie2021a}. Major differences exist w.r.t.\ the following properties:
\begin{itemize}
	\item \textbf{target-based vs.\ target-free}: Most calibration methods use special calibration targets (or environments). In most cases, textured objects with a relatively simple geometry are used, e.g.\ checkerboard patterns, spheres, or cubes. The design of these calibration targets is optimized w.r.t.\ the perception properties of the sensors. A planar target for the simultaneous calibration of cameras, lidar sensors, and radar sensors was developed by \cite{domhof2019a}. In contrast, target-free methods use the unstructured environment of the robot to estimate the calibration parameters. The suitability of these environments for parameter estimation must be ensured. A method for the calibration of radar sensors w.r.t.\ lidar sensors was published by \cite{heng2020a}.
	\item \textbf{online vs.\ offline}: Calibration can be performed during (online) or before/after (offline) the operation of a robot. In mobile robotics, target-based methods are mostly performed offline in the course of a dedicated calibration procedure. Depending on the temporal stability of the calibration, the estimated parameters can differ from the ones during operation of the robot. Online methods, however, estimate (and possibly continuously refine) the calibration parameters. This is especially useful in case of a continuous miscalibration of the sensors, e.g.\ due to thermal influences or mechanical stress. An online calibration method can also be convenient if the relative position of the sensors is often changed, e.g.\ due to a frequent re-arrangement of the sensors (as it is the case at the AIT). An online calibration based on the matching of planes and edges which continuously also tracks the parameter uncertainties was published by \cite{jiao2021a}.
\end{itemize}

In general, the choice of the optimal calibration method strongly depends on the specifics of a multi-sensor system. Thus, it is emphasized, that no definitive statement about the best properties of a calibration can be made.


\section{METHOD}

\subsection{Calibration design}
\label{sec:calibration_design}

The method proposed herein aims to estimate the \textit{extrinsic calibration} of an extereoceptive sensor. The main properties of the proposed calibration are:
\begin{itemize}
	\item The calibration is applicable to any kind of sensor which provides 2D or 3D point clouds of the environment, either directly or indirectly. Lidar sensors, for instance, directly provide point clouds as opposed to stereo cameras or imaging radars which only indirectly provide point clouds through stereo matching and radar target extraction, respectively.
	\item The extrinsic calibration of a sensor is estimated w.r.t.\ the coordinate system of another sensor (\textit{reference sensor}). The point clouds observed by these two sensors must overlap in object space.
	\item The calibration is continuously running during operation of the robot (\textit{on-site calibration}\footnote{sometimes also denoted as \textit{on-the-job calibration} or \textit{self calibration}}), i.e.\ no dedicated calibration procedure or post-processing of data is required. Consequently, we also do not use any special calibration patterns or objects. Instead, the immediate surroundings of the robot, observed as point clouds, are used for calibration.
	\item The extrinsic calibration is estimated from point clouds which are acquired while the robot is static, i.e.\ not moving. This has two main advantages:
	\begin{enumerate}
		\item The path of the robot is not part of the optimization problem. Thus, a possibly erroneous path has no negative influence on the calibration process. Otherwise, the estimated calibration parameters might compensate for these errors due to the well-known correlation of the calibration parameters to the robot's path \cite{glira2019a}.
		\item Time stamps also do not have to be considered in the optimization problem. Thus, erroneous time stamps, e.g.\ due to a slightly incorrect time synchronization of the sensors, have no negative influence on the calibration.
	\end{enumerate}
	\item Depending on the observed scene, a single static position might not be sufficient to estimate the extrinsic calibration with acceptable precision. Thus, the calibration is refined iteratively when the robot reaches a new static position. Thereby, the previous parameter estimates, together with its precision estimates, are used as a priori observations (see next point). The whole calibration process stops once the precision of the calibration parameters are below a user-defined threshold, or in other words, once the calibration is sufficiently accurate.
	\item If a priori observations (measurements) of the 6 calibration parameters (or a subset thereof) exist, they are considered in the estimation process. Such observations can stem e.g.\ from 3D models of the multi-sensor system, from manual measurements (e.g.\ by using a measuring tape), or from a previously performed calibration. These observations are weighted in the least squares optimization according to their observation precision (uncertainty).
	\item It is possible to omit the estimation of individual calibration parameters. This is useful if some of the parameters are known in advance with very high precision and the calibration procedure is not expected to improve these estimates. In practice, this often applies to the translation vector which can be measured with a precision in the sub-millimeter range by other means, e.g.\ by using a total station. In contrast, it is typically rather difficult to directly measure the rotational components of the extrinsic calibration. In this context it should be noted, that incorrect angles can lead to very large displacements in object space, as the effect of angular errors is directly proportional to the range of the observed objects.
\end{itemize}

\begin{figure}[b]
	\centering
	\includegraphics[scale=1]{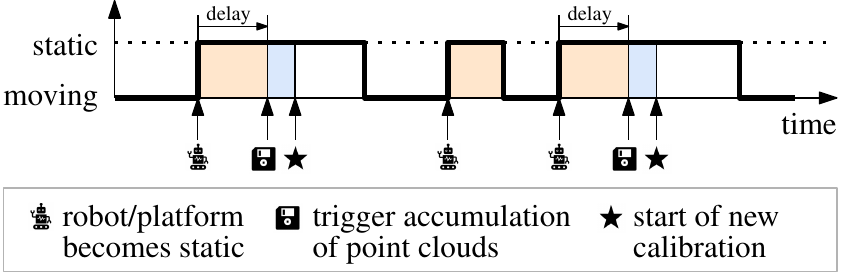}
	\caption{Temporal sequence of the calibration.}
	\label{calibration_site_trigger}
\end{figure}

As prerequisites for the our method we assume that:
\begin{itemize}
	\item approximate values for the 6 calibration parameters are known in advance. This prerequisite stems from the ICP (iterative closest point) algorithm \cite{besl1992a} which is used to estimate the 6 DoF transformation between the two overlapping sensor point clouds. In practice, the rotation angles need to be known typically with a precision of a few degrees and the translation vector with a precision of a few centimeters.
	\item the sensors capture rather dense 3D point clouds of the environment. However, if a sensor provides a 2D point cloud only, i.e.\ a single profile of the environment, the calibration is still applicable, but depending on the observed scene it might be difficult to estimate all 6 parameters of its extrinsic calibration. Consequently, only a subset of the 6 parameters should be estimated in such cases.
\end{itemize}

Finally it should be noted, that the calibration method is \textit{not} applicable to sensors which do \textit{not} observe the object space, e.g.\ navigation sensors like GNSS or IMUs, c.f.\ Fig.~\ref{fig:sensors-troc}.

\subsection{Calibration procedure}
\label{sec:calibration_procedure}

The temporal sequence of the calibration is depicted in Fig.~\ref{calibration_site_trigger}. We distinguish between a static and a moving state of the robot, e.g.\ derived by the current angular and linear velocity of the robot (twist). Each time the robot becomes static, the accumulation of point clouds is triggered after a certain time delay (e.g.\ 2~seconds). The delay should ensure that the robot comes completely at rest before the data acquisition starts. As soon as the accumulation of point clouds is completed, a new calibration is started. This sequence is repeated until the estimated calibration parameters are sufficiently accurate.

\subsection{Calibration estimation}
\label{sec:calibration_estimation}

The extrinsic calibration of a sensor is estimated indirectly through \textit{point cloud matching}\footnote{sometimes also denoted as \textit{point cloud registration}}. More specifically, the point cloud of the sensor to calibrate is matched with the overlapping point cloud of another sensor, denoted as the \textit{reference sensor} throughout this paper. Consequently, the estimated calibration parameters describe the transformation between the two sensor coordinate systems.

\begin{figure}[b]
	\centering
	\includegraphics[scale=0.80]{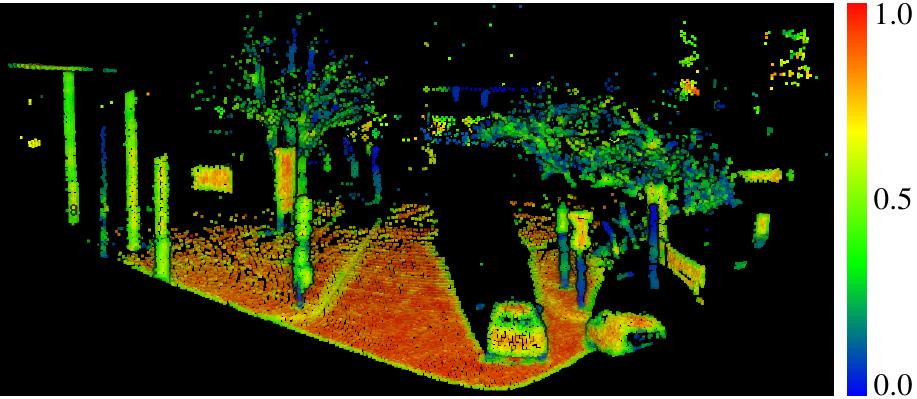}
	\caption{Lidar point cloud colored by planarity. Points with low planarity are filtered out before matching the point clouds.}
	\label{planarity}
\end{figure}

\begin{figure}[thpb]
	\centering
	\includegraphics[width=0.65\linewidth]{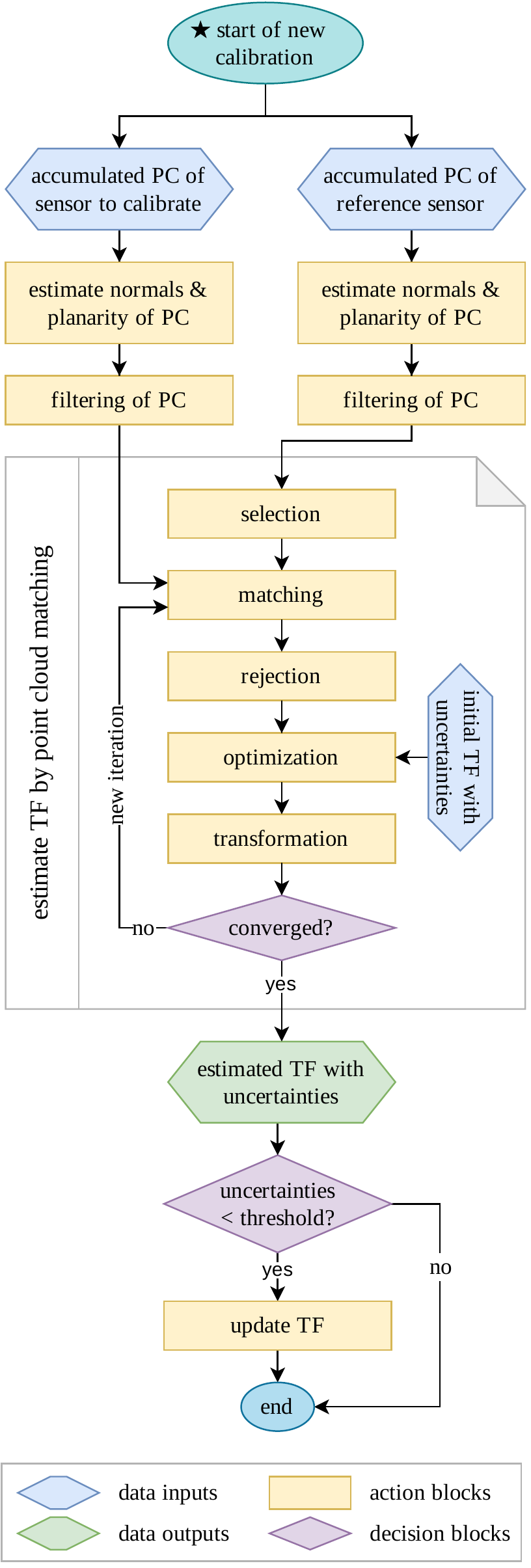}
	\caption{Processing workflow of a single calibration (PC = point cloud, TF = transformation).}
	\label{calibration_site_graph}
\end{figure}

The whole processing workflow of a single calibration is depicted in Fig.~\ref{calibration_site_graph}. The two accumulated sensor point clouds are used as data input, c.f.\ Fig.~\ref{calibration_site_trigger}. First, the normal vector and a corresponding planarity value are estimated for each point. The planarity values range from 0 to 1, where 1 corresponds to a perfect plane \cite{weinmann2014a}. Then, some basic point cloud filtering is carried out. This typically includes a minimum range filter, a maximum range filter, an intensity-based filter, a voxel-based thin out, and a minimum planarity filter. The latter is used to keep only planar areas (e.g.\ roofs, streets, walls) of the point cloud, whereas the non-planar areas (e.g.\ vegetation, edges, corners) are filtered out (Fig.~\ref{planarity}). Additional filters strongly depend on the specifics of a sensor.

We use an ICP-like algorithm for point cloud matching with extended features. It optimizes the alignment of the two sensor point clouds by iteratively minimizing the distances (discrepancies) within the overlap area of these point clouds. This is accomplished by transforming in each iteration the point cloud of the sensor to calibrate, whereas the point cloud of the reference sensor remains fixed. As pointed out in section~\ref{sec:calibration_design}, an approximate initial estimate of the relative alignment of the point clouds is needed. The algorithm can be broken down into five main steps \cite{glira2015a}, c.f. Fig.~\ref{calibration_site_graph}:
\begin{enumerate}
	\item \textbf{selection}: A subset of points is selected in the point cloud of the reference sensor. These points are selected within the overlap area of the two point clouds. We apply the \textit{uniform sampling} selection strategy \cite{glira2015a} which leads to a homogeneous distribution of the selected points in object space.
	\item \textbf{matching}: The corresponding points of the selected subset are searched in the point cloud of the sensor to calibrate. As corresponding point the nearest neighbor is used. The nearest neighbor search is efficiently solved using a k-d tree.
	\item \textbf{rejection}: False correspondences (outliers) are detected and rejected by checking the compatibility of corresponding points. Specifically, we reject correspondences on the basis of their distance and on the basis of the angle between their normals.
	\item \textbf{optimization}: The transformation parameters are estimated for the sensor point cloud by minimizing the distances between corresponding points. A detailed description of this step follows in the next section.
	\item \textbf{transformation}: The sensor point cloud is transformed using the estimated parameters.
\end{enumerate}
Finally, a suitable convergence criterion is tested. If it is not met, the process restarts from the matching step.

Once the ICP algorithm converged, the estimated precision (uncertainty) of the transformation parameters is compared with a user-defined threshold. Consequently, the transformation between the sensors is updated only if the observed scene was well-suited for the parameter estimation.

\subsection{Optimization}
\label{sec:optimization}

The 6 unknown parameters of the extrinsic calibration, specifically the rotation angles $\alpha_x$, $\alpha_y$, $\alpha_z$ and the translation components $t_x$, $t_y$, $t_z$, are estimated in a non-linear weighted least squares adjustment with conditions only (a.k.a.\ \textit{Gauß-Markov} adjustment model); the solution formulas can be found e.g.\ in \cite{mikhail1976a} (chapter 6) or \cite{foerstner2016a} (chapter 4.4). The objective of the adjustment is to minimize the weighted sum of squared residuals:
\begin{equation}
	\Omega = \operatorname*{argmin}_{\alpha_x, \alpha_y, \alpha_z, t_x, t_y, t_z} \left\{ \sum_{i=1}^n p_i r_i^2 \right\}
\end{equation}
where $r_i$ is the residual of the $i$-th observation, $p_i$ the corresponding weight, and $n$ the total number of observations.

We minimize two types of residuals:
\begin{enumerate}
	\item \textbf{Point-to-plane distances between corresponding points}. This observation type is the main element of most contemporary ICP implementations.

	The residual is defined for the $k$-th correspondence as
	\begin{equation}
		r_k = ((R\mathbf{p}_k + \mathbf{t}) - \mathbf{q}_k)^T \mathbf{n}_k
	\end{equation}
	where $\mathbf{p}_k$ and $\mathbf{q}_k$ are the corresponding points of the sensor to calibrate (movable) and the reference sensor (fixed), respectively, $R$ is the rotation matrix composed by the Euler angles $\alpha_x$, $\alpha_y$, $\alpha_z$, $\mathbf{t}$ is the translation vector with its components $t_x$, $t_y$, $t_z$, and $\mathbf{n}_k$ is the normal vector of $\mathbf{q}_k$. We prefer the signed point-to-plane error metric over alternative error metrics (e.g.\ point-to-point) due to its high convergence speed \cite{rusinkiewicz2001a}, its straight-forward mathematical formulation, and the fact that corresponding points only need to belong to the same plane\footnote{In contrast, when using the point-to-point error metric, corresponding points need to be exactly identical. However, such correspondences are typically very unlikely, as point clouds randomly sample the object space.} \cite{glira2015a}.

	The weight of these residuals is defined according to the theory of least squares \cite{mikhail1976a} (chapter 3.3) as
	\begin{equation}
		p_k = 1 / \sigma_d^2
	\end{equation}
	where we propose to estimate $\sigma_d$ from all a priori point-to-plane distances as follows:
	\begin{equation}
		\sigma_d = 1.4826 \cdot \textrm{mad}
	\end{equation}
	Thereby, $\textrm{mad}$ denotes the \textit{median of absolute differences} (w.r.t. the median) \cite{hampel1974a} of the point-to-plane distances. It is commonly used as a robust estimator for the standard deviation of a set of random variables which is generally assumed to be normally distributed, but still contaminated by a small number of outliers.

	\item \textbf{Differences to initial values of transformation parameters}. This observation type is crucial for our method as it allows (from the second calibration site onwards) to appropriately transfer the estimates from a previoulsy performed calibration. Additionally, these observations can be used (at the first calibration site) to consider any other a priori estimates of the parameters, e.g.\ estimates from a 3D model or from manual measurements (c.f.\ section~\ref{sec:calibration_design}).

	The residuals are defined for each transformation parameter as follows:
	\begin{align}
		v_{\alpha_x} &= \overline{\alpha}_x - \alpha_x \nonumber \\
		v_{\alpha_y} &= \overline{\alpha}_y - \alpha_y \nonumber \\
		v_{\alpha_z} &= \overline{\alpha}_z - \alpha_z \\
		v_{t_x} &= \overline{t}_x - t_x \nonumber \\
		v_{t_y} &= \overline{t}_y - t_y \nonumber \\
		v_{t_z} &= \overline{t}_z - t_z \nonumber
	\end{align}
	where $\overline{\alpha}_x$, $\overline{\alpha}_y$, $\overline{\alpha}_z$, $\overline{t}_x$, $\overline{t}_y$, $\overline{t}_z$ are the observed initial values and $\alpha_x$, $\alpha_y$, $\alpha_z$, $t_x$, $t_y$, $t_z$ are the estimated unknown parameters.

	These residuals are weighted by considering their uncertainty estimates, e.g.\ for $\alpha_x$:
	\begin{equation}
		p_{\alpha_x} = 1 / \sigma_{\alpha_x}^2
	\end{equation}
	where $\sigma_{\alpha_x}$ is the precision of the observed value. If the initial values stem from a previous calibration, their squared precision (i.e.\ their variance) is given by the diagonal of the a posteriori covariance matrix of the unknown parameters, i.e.\ $\textrm{diag}(C_{\hat{x}\hat{x}})$.
\end{enumerate}

\subsection{Implementation}
\label{sec:implementation}

The calibration is implemented as package with a single node for ROS2 (robot operating system). It is written mainly in python. The Point Data Abstraction Library (PDAL\footnote{pdal.io}) is used for pre-processing of the accumulated point clouds. The ICP-like point cloud matching algorithm is mainly based upon the following libraries: numpy, pandas, scipy, and lmfit\footnote{numpy.org, pandas.pydata.org, scipy.org, lmfit.github.io}. We named this algorithm ``simpleICP'' and published it as open source on Github and PyPi under the MIT license \cite{glira2022b}. Development tests have been made mainly with the simulation environment Webots\footnote{cyberbotics.com}. When calibrating multiple sensors at the same time, an independent node is used for each sensor pair, e.g.\ sensor1-to-sensor2, sensor1-to-sensor3, sensor1-to-sensor4, etc.. Currently, the full processing of the pipeline as depicted in Fig.~\ref{calibration_site_graph} takes approximately 3--5 seconds.


\section{EXPERIMENTAL RESULTS}
\label{sec:experimental_results}

\begin{figure}[thpb]
	\centering
	\includegraphics[scale=1]{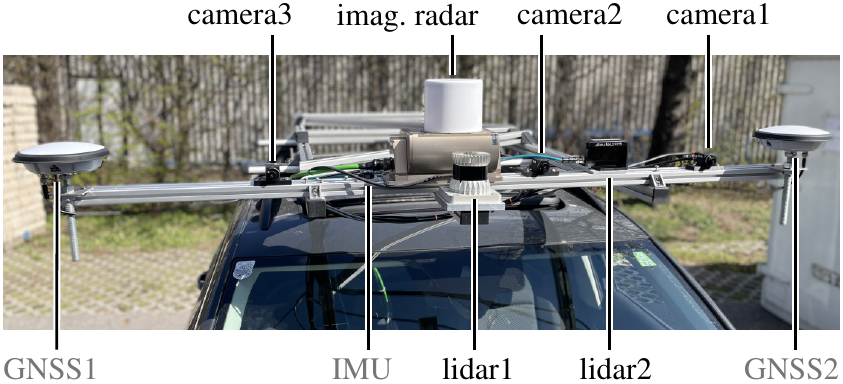}
	\caption{The proposed method was applied to estimate the extrinsic calibration of 3 cameras, 2 lidars, and an imaging radar sensor. (The calibration of the GNSS and IMU navigation sensors, however, is not within the scope of this method.)}
	\label{fig:sensors-troc}
\end{figure}

The proposed calibration method was applied to estimate the extrinsic calibration of the sensors depicted in Fig.~\ref{fig:sensors-troc}. The sensor rig was mounted on a car and is composed by an \textit{Ouster OS1-64} (lidar1), a \textit{Blickfeld Cube 1} (lidar2), 3 \textit{ptgrey} cameras with 1/3'' CMOS sensor and a resolution of 1.3\,Mpx (camera1/2/3), and the \textit{Indurad iSDR-p} (imaging radar).

\begin{figure}[thpb]
	\centering
	\includegraphics[scale=1]{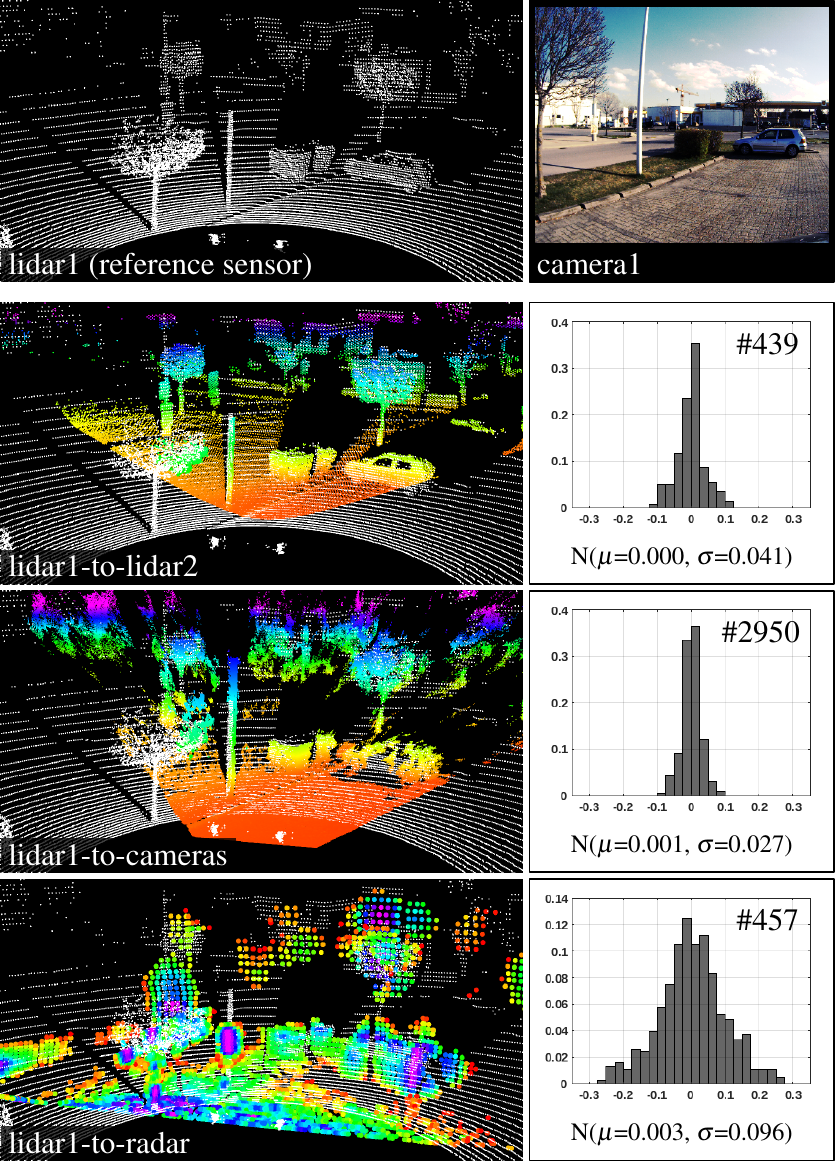}
	\caption{Data belonging to calibration site 7, c.f.\ Fig.~\ref{fig:sensorscalibration_ait-troc}. First row: point cloud of the reference sensor lidar1 and a corresponding image of the site. Rows 2--4: point clouds of the sensors to calibrate (left) and histograms of the residual point-to-plane distances (right). (The point clouds of lidar2 and of the cameras are colored by height, whereas the radar point cloud is colored by measured intensity.)}
	\label{fig:stats_sensors}
\end{figure}

As pointed out in \ref{sec:calibration_design}, each sensor must provide a point cloud in order to apply our calibration method. Both lidar sensors directly provide point clouds. The stereo matching point cloud is derived from the camera images using the algorithm described in \cite{humenberger2010a}. The imaging radar provides so-called range-amplitude maps; a point cloud is generated by extracting radar targets from these maps \cite{glira2022a}. In Fig.~\ref{fig:stats_sensors} the point clouds provided by these sensors are visualized for a single calibration site.

\begin{figure*}[thpb]
	\centering
	\includegraphics[scale=1]{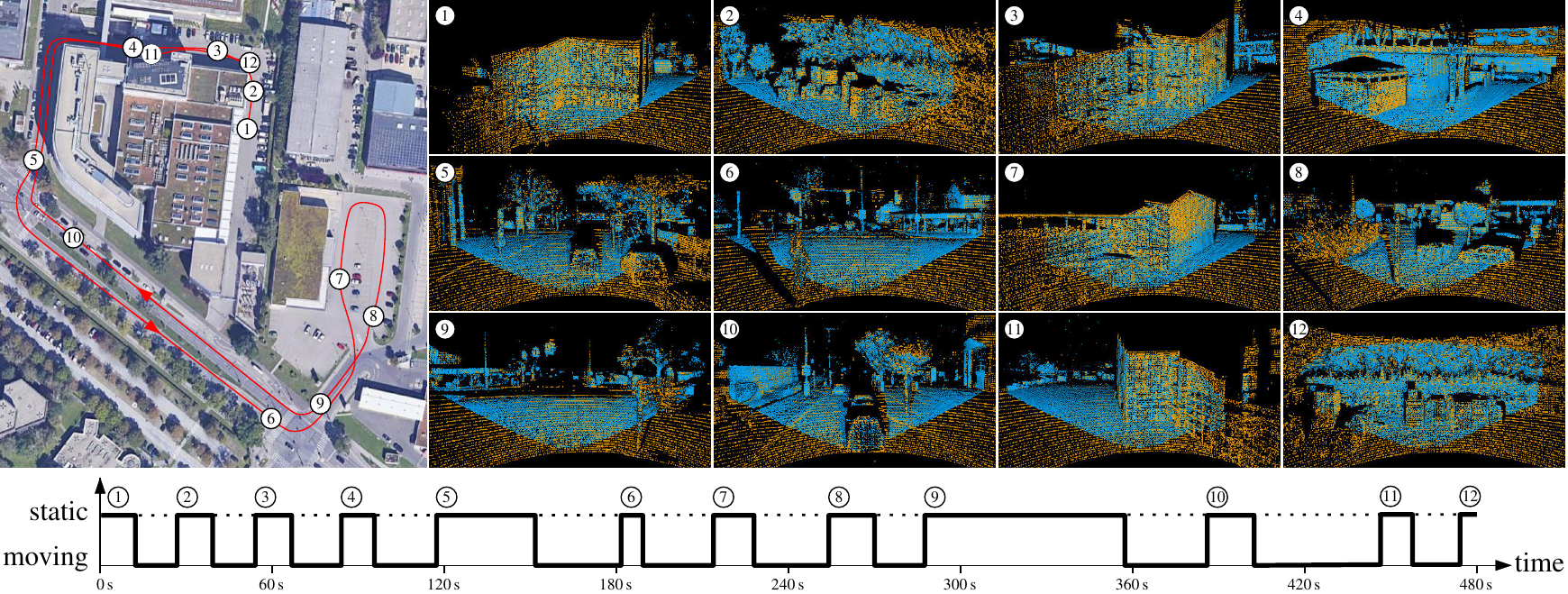}
	\caption{The multi-sensor system depicted in Fig.~\ref{fig:sensors-troc} was calibrated during an 8 minute drive at 12 different calibration sites.}
	\label{fig:sensorscalibration_ait-troc}
\end{figure*}

The sensors have been calibrated during an 8 minute drive near the AIT headquarter (Fig.~\ref{fig:sensorscalibration_ait-troc}). In this time period, the car stopped 12 times. Accordingly, the calibration was triggered at 12 different calibration sites. In order to distinguish between the static and moving state of the car, GNSS measurements have been processed in a relatively simple Kalman filter. Fig.~\ref{fig:sensorscalibration_ait-troc} shows the position of the calibration sites, as well as the corresponding point clouds as collected by the sensors lidar1 and lidar2.

It is recommended to use the sensor with the highest measurement accuracy and a large field of view (FOV) as \textit{reference sensor}. Ideally, the point cloud of this sensor shares a large overlap area with the point clouds of the other sensors. Consequently, we have chosen the sensor lidar1 as reference for our multi-sensor system. It has the highest measurement accuracy and the largest FOV due to its \ang{360} rotating antenna. The extrinsic calibration of the other sensors was estimated in parallel by 3 independent calibration procedures: lidar1-to-lidar2, lidar1-to-cameras, lidar1-to-radar. Fig.~\ref{fig:stats_sensors} shows the point clouds and the histograms of the residual point-to-plane distances for a single calibration site. As can be seen, the mean of the residuals is very close to zero in all 3 cases. The standard deviation, however, differs for each sensor combination as it mainly depends on the measurement accuracy of the sensors and on the selection of correspondences, specifically the minimum planarity value used for point cloud filtering.

\begin{figure}[h]
	\centering
	\includegraphics[width=0.95\linewidth]{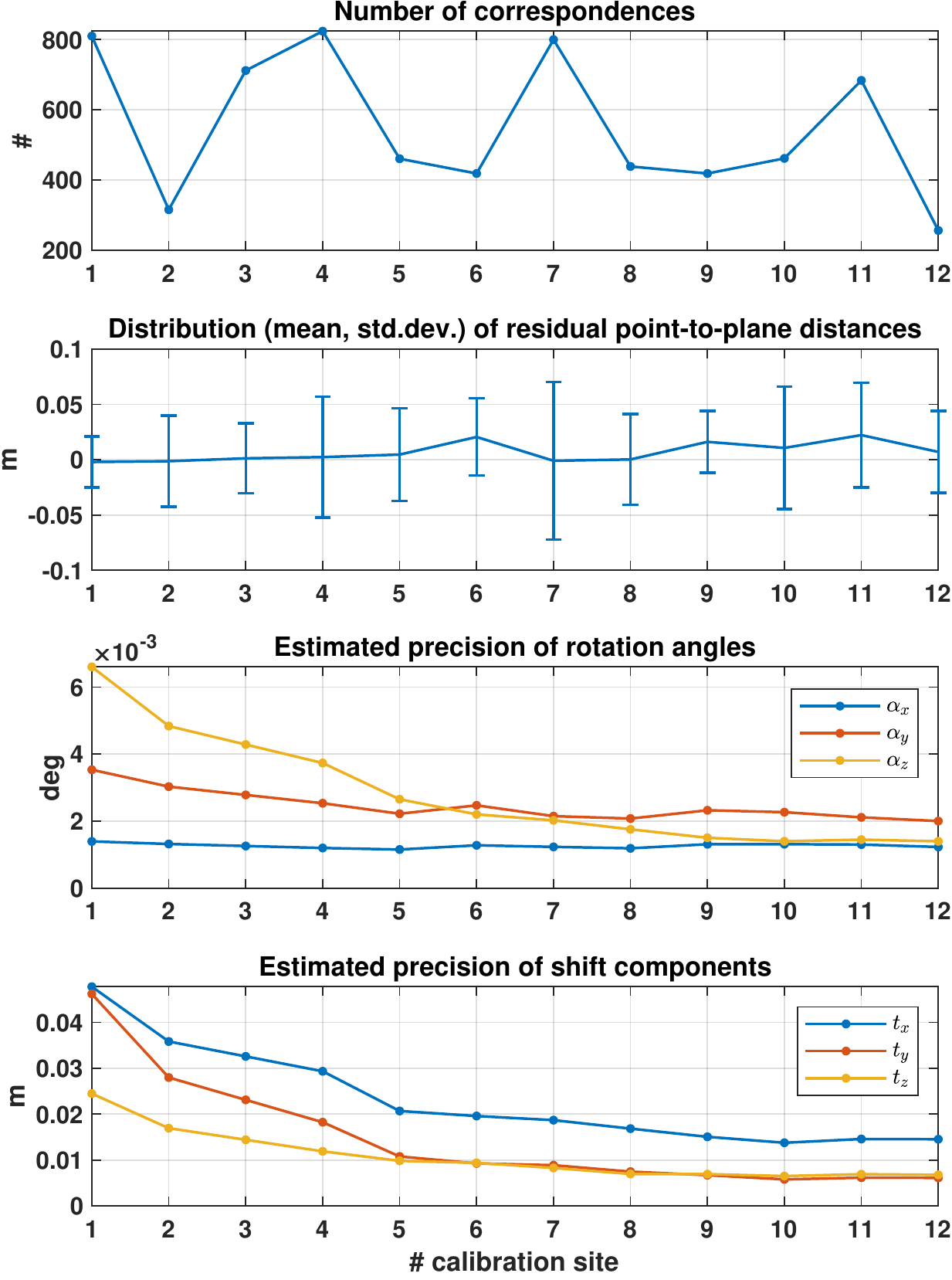}
	\caption{Temporal development of some calibration quality indicators for the sensor combination lidar1-to-lidar2.}
	\label{fig:iteration_stats}
\end{figure}

Fig.\ref{fig:iteration_stats} shows, specifically for the sensor combination lidar1-to-lidar2, the temporal development of some calibration quality indicators. The first graph shows the number of correspondences used in the optimization. This number mainly depends on the amount of planar areas in the scene, c.f.\ Fig.~\ref{fig:sensorscalibration_ait-troc}. The second graph shows the distribution of the residual point-to-plane distances. The standard deviation, again, strongly depends on the characteristics of the observed scene. On the contrary, the mean value shows a systematic behavior: it is very close to zero in the first few iterations but gets slightly larger in magnitude later. This can be explained through the iterative refinement character of our calibration method: relatively seen, the influence of the point-to-plane residuals (residual type 1, c.f.\ section~\ref{sec:optimization}) on the parameter estimation is highest at the first calibration site. Subsequently, however, the relative influence (i.e.\ their weights) of the second type of residuals, the direct observation of transformation parameters from previous calibrations, continuously grows. Finally, the third and fourth graphs show the estimated precision of the transformation parameters as given by the a posteriori covariance matrix of the unknown parameters $C_{\hat{x}\hat{x}}$. Here, one can observe that their precision is improving over time which can be regarded as one of the main benefits of our method.


\section{SUMMARY AND OUTLOOK}
\label{sec:summary_and_outlook}

We proposed in this work a new method for the extrinsic calibration which:
\begin{itemize}
	\item does not use any special calibration targets (target-free)
	\item is widely sensor agnostic
	\item can be applied simultaneously to multiple sensors
	\item continuously improves the calibration parameter estimates over time
\end{itemize}

Our future work will concentrate on:
\begin{itemize}
	\item the automatic removal of dynamic objects (like cars, persons, etc.) from the collected point clouds
	\item the calibration of sensors which provide 2D point clouds only, i.e.\ a profile of the environment
	\item better understand the long-time behavior of the estimated precision of the calibration parameters, e.g.\ for an operation time of several hours
\end{itemize}

In addition, we plan to publish the code as open source package for ROS2.





\bibliographystyle{plain}
\bibliography{literature.bib}

\begin{thebibliography}{10}

\bibitem{besl1992a}
Paul~J Besl and Neil~D McKay.
\newblock Method for registration of 3-d shapes.
\newblock In {\em Robotics-DL tentative}, pages 586--606. International Society
  for Optics and Photonics, 1992.

\bibitem{domhof2019a}
Joris Domhof, Julian~F.P. Kooij, and Dariu~M. Gavrila.
\newblock An extrinsic calibration tool for radar, camera and lidar.
\newblock In {\em 2019 International Conference on Robotics and Automation
  (ICRA)}, pages 8107--8113, 2019.

\bibitem{foerstner2016a}
Wolfgang F{\"o}rstner and Bernhard Wrobel.
\newblock {\em Photogrammetric Computer Vision -- Statistics, Geometry,
  Orientation and Reconstruction}.
\newblock Springer, 2016.

\bibitem{glira2022b}
Philipp Glira.
\newblock simple{ICP}: Implementations of a rather simple version of the
  iterative closest point algorithm in various languages.
\newblock https://github.com/pglira/simpleICP, April 2022.

\bibitem{glira2015a}
Philipp Glira, Norbert Pfeifer, Christian Briese, and Camillo Ressl.
\newblock A correspondence framework for {ALS} strip adjustments based on
  variants of the {ICP} algorithm.
\newblock {\em PFG Photogrammetrie, Fernerkundung, Geoinformation},
  2015(4):275--289, 08 2015.

\bibitem{glira2019a}
Philipp Glira, Norbert Pfeifer, and Gottfried Mandlburger.
\newblock Hybrid orientation of airborne lidar point clouds and aerial images.
\newblock {\em ISPRS Annals of Photogrammetry, Remote Sensing \& Spatial
  Information Sciences}, 4, 2019.

\bibitem{glira2022a}
Philipp Glira, Christoph Weidinger, Thomas Kadiofsky, Wolfgang Pointner,
  Katharina {\"O}lsb{\"o}ck, Christian Zinner, and Masrur Doostdar.
\newblock 3d mobile mapping of the environment using imaging radar sensors.
\newblock {\em Proceedings of RadaConf2022}, 2022.

\bibitem{hampel1974a}
Frank~R Hampel.
\newblock The influence curve and its role in robust estimation.
\newblock {\em Journal of the American Statistical Association},
  69(346):383--393, 1974.

\bibitem{heng2020a}
Lionel Heng.
\newblock Automatic targetless extrinsic calibration of multiple 3d lidars and
  radars.
\newblock In {\em 2020 IEEE/RSJ International Conference on Intelligent Robots
  and Systems (IROS)}, pages 10669--10675, 2020.

\bibitem{humenberger2010a}
Martin Humenberger, Christian Zinner, Michael Weber, Wilfried Kubinger, and
  Markus Vincze.
\newblock A fast stereo matching algorithm suitable for embedded real-time
  systems.
\newblock {\em Computer Vision and Image Understanding}, 114(11):1180--1202,
  2010.

\bibitem{jiao2021a}
Jianhao Jiao, Haoyang Ye, Yilong Zhu, and Ming Liu.
\newblock Robust odometry and mapping for multi-lidar systems with online
  extrinsic calibration.
\newblock {\em IEEE Transactions on Robotics}, 2021.

\bibitem{mikhail1976a}
E.M. Mikhail and F.E. Ackermann.
\newblock {\em Observations and least squares}.
\newblock University Press of America, 1976.

\bibitem{nie2021a}
Jiwei Nie, Feng Pan, Dingyu Xue, and Ling Luo.
\newblock A survey of extrinsic parameters calibration techniques for
  autonomous devices.
\newblock In {\em 2021 33rd Chinese Control and Decision Conference (CCDC)},
  pages 3543--3548, 2021.

\bibitem{rusinkiewicz2001a}
Szymon Rusinkiewicz and Marc Levoy.
\newblock Efficient variants of the {ICP} algorithm.
\newblock In {\em 3-D Digital Imaging and Modeling, 2001. Proceedings. Third
  International Conference on}, pages 145--152, Quebec City, Canada, 2001.
  IEEE.

\bibitem{siegwart2011a}
Roland Siegwart, Illah~Reza Nourbakhsh, and Davide Scaramuzza.
\newblock {\em Introduction to autonomous mobile robots}.
\newblock MIT press, 2011.

\bibitem{weinmann2014a}
Martin Weinmann, Boris Jutzi, and Cl{\'e}ment Mallet.
\newblock Semantic 3d scene interpretation: A framework combining optimal
  neighborhood size selection with relevant features.
\newblock {\em ISPRS Annals of the Photogrammetry, Remote Sensing and Spatial
  Information Sciences}, 2(3):181, 2014.

\end{thebibliography}

\end{document}